\documentclass[journal]{IEEEtran}

\ifCLASSINFOpdf
\else
\fi

\usepackage{amsmath}
\usepackage{orcidlink}
\usepackage{multirow}
\usepackage{booktabs}
\usepackage{tabularx}
\usepackage{graphicx}
\usepackage{caption}
\usepackage{amsthm}
\usepackage{amsfonts}   % 基本数学字体（包含 \mathbb）
\usepackage{amssymb}    % 扩展数学符号（推荐，包含 amsfonts 的功能）
\usepackage{colortbl,xcolor} % 放到导言区
\definecolor{lightred}{RGB}{255,230,230}
\definecolor{lightblue}{RGB}{230,240,255}

\newtheorem{theorem}{Theorem}

\makeatletter
    \def\@cite#1#2{\textsuperscript{[{#1\if@tempswa , #2\fi}]}}
\makeatother

\begin{document}
\title{AR-KAN: Autoregressive-Weight-Enhanced Kolmogorov-Arnold Network for Time Series Forecasting}

\author{
        Chen~Zeng,
        Tiehang~Xu,
    and~Qiao~Wang\orcidlink{0000-0002-5271-0472},~\IEEEmembership{Senior Member,~IEEE}%
 
\thanks{Both  C. Zeng and T. Xu was with the School of Information Science and Engineering, Southeast University, Nanjing, China (email:
chenzeng@seu.edu.cn,
220250920@seu.edu.cn). }
\thanks {Q. Wang was with both the School of Information Science and Engineering and the School of Economics and Management, Southeast University, Nanjing, China (Corresponding Author, email: qiaowang@seu.edu.cn).}}

\maketitle

\begin{abstract}
Traditional neural networks struggle to capture the spectral structure of complex signals. Fourier neural networks (FNNs) attempt to address this by embedding Fourier series components, yet many real-world signals are almost-periodic with non-commensurate frequencies, posing additional challenges. Building on prior work \cite{cao-2024} showing that ARIMA outperforms large language models (LLMs) for time series forecasting, we extend the comparison to neural predictors and find that ARIMA still maintains a clear advantage. Inspired by this finding, we propose the Autoregressive-Weight-Enhanced Kolmogorov-Arnold Network (AR-KAN). Based in the Universal Myopic Mapping Theorem, it integrates a pre-trained AR module for temporal memory with a KAN for nonlinear representation. We prove that the AR module preserves essential temporal features while reducing redundancy, and that the upper bound of the approximation error for AR-KAN is smaller than that for KAN in a probabilistic sense. Experimental results also demonstrate that AR-KAN delivers exceptional performance compared to existing models, both on synthetic almost-periodic functions and real-world datasets. These results highlight AR-KAN as a robust and effective framework for time series forecasting. Our code is available at \textit{\url{https://github.com/ChenZeng001/AR-KAN}}.
\end{abstract}


\IEEEpeerreviewmaketitle

\section{Introduction}

Time series forecasting is a fundamental task in signal processing\cite{box2015}\cite{wienerGHA}, statistics\cite{statistics}, and numerous applied fields, including economics\cite{economics}, meteorology\cite{meteorology}, and healthcare\cite{healthcare}. Among classical approaches, the Autoregressive Integrated Moving Average (ARIMA) model\cite{arima} stands out as one of the most influential and widely adopted methods, because it integrates autoregression, differencing, and moving average elements to provide a comprehensible and effective approach for handling practical time series data, even when the time series is non-stationary.

Apart from the aforementioned statistics or Fourier analysis-based methods, neural networks have been utilized in time series forecasting for many years\cite{lim-2021}, with the goal of enabling the modeling of complex nonlinear dependencies. Architectures such as Multi-Layer Perceptrons (MLPs)\cite{mlp} and Recurrent Neural Networks (RNNs)\cite{rnn}, particularly Long Short-Term Memory (LSTM) networks\cite{lstm}, have been widely studied. In recent years, Transformer-based models\cite{transformer1}\cite{transformer2}\cite{transformer3} have gained popularity due to their self-attention mechanism and parallel processing capabilities. Alongside these developments, Convolutional Neural Networks (CNNs)\cite{cnn} have also been successfully adapted for temporal data. Notably, Temporal Convolutional Networks (TCNs)\cite{tcn} have emerged as a highly effective alternative to traditional recurrent architectures. More recently, Kolmogorov-Arnold Networks (KANs)\cite{kan}\cite{kan2} have been introduced as a novel architecture with high expressivity and flexible modeling of nonlinear mappings. In parallel, the rapid progress of large language models (LLMs) has led to approaches such as LLMTime\cite{llmtime} and Time-LLM\cite{time-LLM}, which adapt pretrained language models to temporal tasks by leveraging their strong generalization and sequence modeling capabilities.

In the context of neural forecasting, a specialized research  focuses on spectral analysis through specific networks, such as Fourier Neural Networks (FNNs)\cite{fnn1}. These models incorporate Fourier series to enhance spectral modeling\cite{fnn2}. Representative examples include TimesNet\cite{timesnet}, the Fourier Neural Operator (FNO)\cite{fno} and the Fourier Analysis Network (FAN)\cite{fan}, which have been applied to physics-informed learning, partial differential equation solving, and time series prediction.

Nevertheless, these  neural network models grounded in representation by Fourier series may overlook a key theoretical constraint: the additive combination of periodic elements does not necessarily result in a periodic function\cite{almost-periodic1}\cite{almost-periodic2}. Throughout history, this important topic prompted N. Wiener to create the renowned Generalized Harmonic Analysis (GHA) theory, which works alongside the spectral analysis of time series. When the constituent frequencies are incommensurable, the resulting signal is almost-periodic\cite{almost-periodic3}, meaning that it exhibits recurrence without strict periodicity. Empirical studies show that for such signals, even advanced neural models, including FNNs, are often outperformed by classical ARIMA\cite{almost-periodic-arima1}\cite{almost-periodic-arima2} and an evaluation could be referred as to our recent work \cite{cao-2024}.

To address this, we propose AR-KAN, a hybrid model that integrates the strengths of traditional and modern approaches. Based on the Universal Myopic Mapping Theorem \cite{ummt1,ummt2}, AR-KAN employs a KAN as the static nonlinear component while introducing temporal memory through a pre-trained autoregressive (AR) module. This design enables AR-KAN to combine the adaptability and expressiveness of KANs with the strong spectral bias inherent in traditional AR models. Furthermore, it can be proven that when handling time series forecasting tasks, the AR module effectively eliminates redundancy while retaining the maximal amount of useful information. We also establish the upper bounds of the approximation errors for both models under simple structures, demonstrating that AR-KAN achieves a smaller error bound than standard KAN in a probabilistic sense.

Extensive experiments demonstrate the effectiveness of AR-KAN. Whether evaluated on synthetic almost-periodic functions or real-world datasets (including Rdatasets\cite{rdatasets}, M3\cite{m3}, and M4\cite{m4}), AR-KAN consistently outperforms other baseline models. Notably, in the experiments involving almost-periodic functions, many models perform even worse than the traditional ARIMA model. This observation underscores that modern neural networks indeed struggle to inherently capture complex spectral properties, reaffirming the value of traditional autoregressive priors. Furthermore, ablation studies also validate the necessity and effectiveness of each proposed component within our architecture.

Our main contributions are four-fold:
\begin{itemize}
\item We propose AR-KAN, a hybrid forecasting model grounded in the Universal Myopic Mapping Theorem. It successfully integrates the temporal memory and strong spectral bias of traditional autoregressive models with the flexible nonlinear expressiveness of KANs.

\item We theoretically prove that the pre-trained AR module eliminates redundancy while preserving maximal useful temporal information. Additionally, We establish the upper bounds of the approximation errors for both models under simple structures, demonstrating that AR-KAN achieves a smaller error bound than standard KAN in a probabilistic sense.

\item We address a critical limitation of existing neural networks in handling almost-periodic signals. By incorporating autoregressive priors, AR-KAN effectively captures complex spectral properties with incommensurable frequencies that modern deep learning models often struggle to learn.

\item We extensively evaluate AR-KAN on synthetic almost-periodic functions and real-world benchmarks (Rdatasets, M3, and M4). Empirical results and thorough ablation studies consistently demonstrate its superior performance and validate the effectiveness of each proposed component.
\end{itemize}

\section{Background}\label{Background}

\subsection{Time Series Forecasting and ARIMA}

Time series forecasting aims to predict a sequence based on its past observations. Formally, given a univariate time series $\{x_n\}_{n=1}^T$, the forecasting problem involves learning a mapping $\mathcal{F}$ such that:

\begin{equation}
    \hat{x}_{n+h} = \mathcal{F}(x_n, x_{n-1}, \dots, x_{n-p+1}),
\end{equation}

where $\hat{x}_{n+h}$ denotes the forecast for $h$-steps ahead ($h=1$ in this paper), and $p$ is the order of historical dependence. This formulation can be extended to multivariate or probabilistic settings, but the central challenge remains: capturing the underlying temporal dynamics, dependencies, and possibly noise in the observed data.

A classical and widely used model for time series forecasting is ARIMA. ARIMA is particularly effective for stationary or differenced stationary processes. The general form of an ARIMA($p$, $d$, $q$) model is given by:

\begin{equation}
    \Phi(B)(1 - B)^d x_n = \Theta(B)\epsilon_n,
\end{equation}
where:
\begin{itemize}
    \item $B$ is the backshift operator, i.e., $B^k x_n = x_{n-k}$,
    \item $\Phi(B) = 1 - a_1 B - \dots - a_{p} B^{p}$ is the autoregressive (AR) polynomial of order $p$,
    \item $\Theta(B) = 1 + b_1 B + \dots + b_{q} B^{q}$ is the moving average (MA) polynomial of order $q$,
    \item $d$ is the degree of differencing to ensure stationarity,
    \item $\epsilon_n$ is assumed to be white noise: $\epsilon_t \sim \mathcal{N}(0, \sigma^2)$.
\end{itemize}

The integration component $(1 - B)^d$ transforms non-stationary series into stationary ones by differencing. The ARIMA model captures linear temporal dependencies and is known for its statistical interpretability and relatively low computational cost. Despite its simplicity, ARIMA remains a strong baseline in many practical applications, especially when the underlying signal exhibits regular, stationary behavior.

\subsection{MLP and KAN}

MLP is one of the most fundamental architectures in neural networks. An MLP consists of multiple layers of affine transformations followed by pointwise nonlinear activations. Given an input $x \in \mathbb{R}^d$, an $L$-layer MLP computes:

\begin{equation}
    f_{\text{MLP}}(x) = W^{(L)} \sigma_{L-1} \cdots \sigma_1 \left( W^{(1)} x + b^{(1)} \right) + b^{(L)},
\end{equation}
where $W^{(\ell)}$, $b^{(\ell)}$ are learnable parameters, and $\sigma_\ell$ denotes the nonlinear activation at layer $\ell$. 

\begin{figure*}[htbp]
    \centering
    \includegraphics[width=0.7\linewidth]{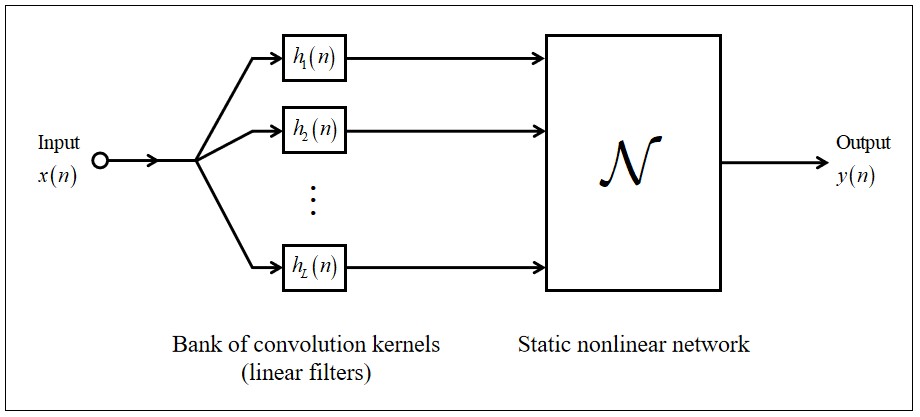}
    \caption{Universal Myopic Mapping Theorem.}
    \label{img-theorem}
\end{figure*}

\begin{figure*}[htbp]
            \centering
            \includegraphics[width=0.7\linewidth]{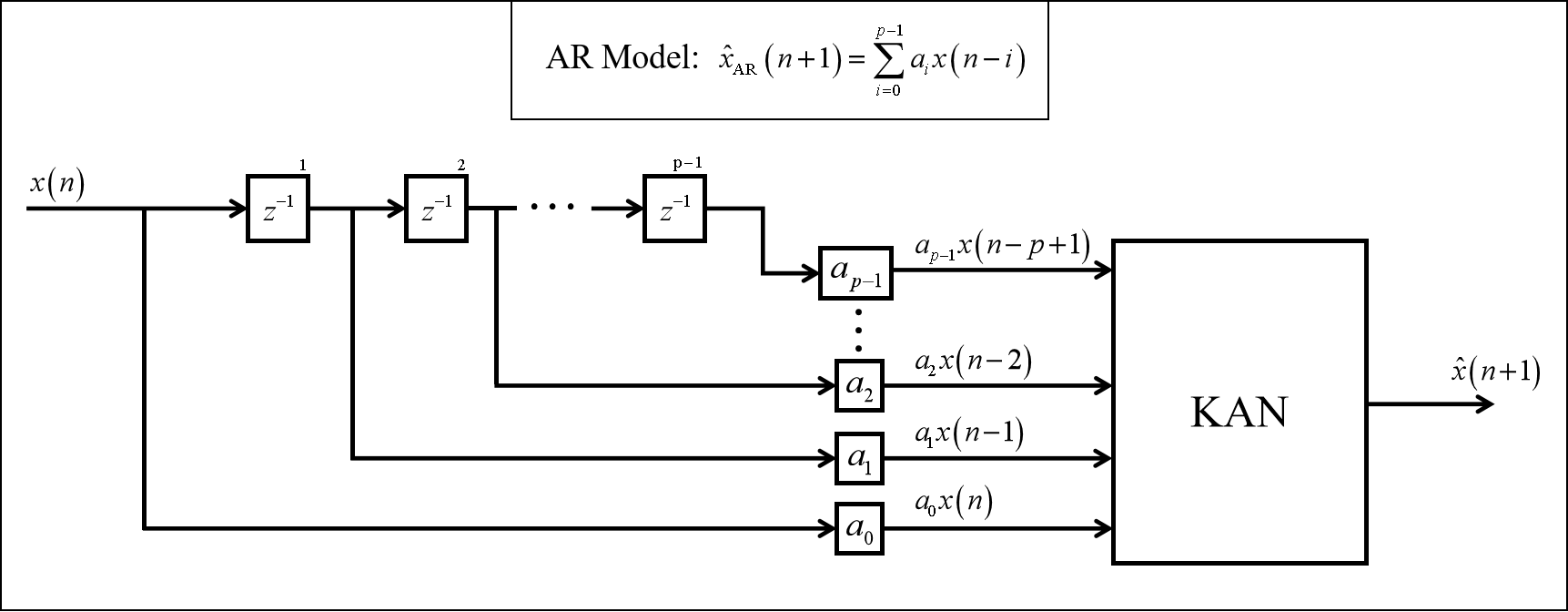}
            \caption{Model Structure of AR-KAN.}
            \label{img-arkan}
\end{figure*}

However, MLPs exhibit a well-known spectral bias\cite{mlp-spectral-bias}, meaning they tend to learn low-frequency components of the target function earlier and more accurately than high-frequency components. While this inductive bias can be beneficial in some applications, it limits the ability of MLPs to capture fine-grained or oscillatory patterns in data.

To overcome the limited expressiveness of fixed activation functions in traditional MLPs, KANs have been proposed as a more flexible and interpretable alternative. KANs are inspired by the Kolmogorov–Arnold representation theorem\cite{kolmogorov}, which states that any multivariate continuous function $f: [0,1]^d \rightarrow \mathbb{R}$ can be expressed as a finite composition of univariate continuous functions:

\begin{equation}
    f(x_1, \dots, x_d) = \sum_{q=1}^{2d+1} \phi_q\left( \sum_{i=1}^d \psi_{qi}(x_i) \right),
\end{equation}
where $\phi_q$ and $\psi_{qi}$ are univariate continuous functions. Inspired by this constructive result, KANs replace the fixed nonlinear activations in MLPs with learnable univariate functions, typically represented by splines.

Given an input $x \in \mathbb{R}^d$, an $L$-layer KAN computes:

\begin{equation}
    f_{\text{KAN}}(x) = \Phi^{(L)} \Psi^{(L-1)} \cdots \Psi^{(1)}(x),
\end{equation}
where each layer $\Psi^{(\ell)}: \mathbb{R}^{d_\ell} \rightarrow \mathbb{R}^{d_{\ell+1}}$ is defined by:

\begin{equation}
    [\Psi^{(\ell)}(x)]_j = \sum_{i=1}^{d_\ell} w_{ij}^{(\ell)} \cdot \psi_{ij}^{(\ell)}(x_i),
\end{equation}
and $\Phi^{(L)}$ denotes the final output transformation, typically of the same form. Here, each $\psi_{ij}^{(\ell)}$ is a learnable univariate function, often implemented using splines, and $w_{ij}^{(\ell)}$ are learnable scalar weights.

Unlike MLPs, KANs do not exhibit a low-frequency spectral bias~\cite{kan-spectral-bias}. This enables them to capture high-frequency and oscillatory components more effectively, making them well suited for modeling time series with rich spectral structures.

However, this advantage can also introduce challenges. Without a low-frequency bias, KANs tend to be more sensitive to high-frequency noise\cite{kan-noise} and may have difficulty learning functions with limited regularity\cite{kan-regular}. In such cases, the model may overfit to spurious variations or become unstable during training.

Nevertheless, in most real-world time series, especially those with structured periodicity, seasonal trends, or non-stationary high-frequency patterns, this characteristic is beneficial. The ability of KANs to model a broad spectrum of frequency behaviors often leads to better performance compared to MLPs.

\section{AR-KAN}\label{AR-KAN}

AR-KAN is derived from the Universal Myopic Mapping Theorem. Therefore, in this section, we first introduce the Universal Myopic Mapping Theorem, then followed by a detailed explanation of the AR-KAN model architecture.

\subsection{Universal Myopic Mapping Theorem}

The Universal Myopic Mapping Theorem~\cite{ummt1}\cite{ummt2} provides a powerful theoretical guarantee for modeling dynamic systems using shallow, feedforward structures. Specifically, it states that any shift-invariant and myopic dynamical map can be uniformly approximated arbitrarily well by a two-stage architecture: a bank of linear filters followed by a static nonlinear mapping, as shown in Fig.~\ref{img-theorem}.

\begin{theorem}[Universal Myopic Mapping Theorem~\cite{ummt1}\cite{ummt2}]
Let $\mathcal{M}$ be a shift-invariant and myopic dynamical system that maps a real-valued time series $\{x_n\}_{n \in \mathbb{Z}}$ to outputs $\{y_n\}$ via a causal and bounded operator. Then, for any $\varepsilon > 0$, there exists a finite collection of linear filters $\{h_i\}_{i=1}^N$ and a continuous static nonlinear function $f_\theta : \mathbb{R}^N \to \mathbb{R}$ such that the approximation
\[
    y_n \approx f_\theta\left( (h_1 * x)_n, (h_2 * x)_n, \dots, (h_N * x)_n \right)
\]
satisfies
\[
    \sup_n \left| y_n - f_\theta\left( (h_1 * x)_n, \dots, (h_N * x)_n \right) \right| < \varepsilon,
\]
where $*$ denotes convolution and $(h_i * x)_n = \sum_{\tau} h_i(\tau) x_{n - \tau}$.
\end{theorem}

This theorem establishes that it is theoretically sufficient to model a wide class of dynamical systems using a finite bank of linear filters followed by a nonlinear function, without requiring recurrent or deep sequential architectures. The key property of myopia means that each output depends only on a bounded past history, and shift-invariance ensures time-homogeneity.

\subsection{Model Structure of AR-KAN}

Inspired by the Universal Myopic Mapping Theorem, we design the AR-KAN as a two-stage architecture composed of a data-driven memory module and a static nonlinear mapping, as illustrated in Fig.~\ref{img-arkan}. The static nonlinear network is implemented using a KAN, which has been discussed in Section \ref{Background} to possess stronger spectral modeling capabilities than traditional MLPs, particularly for high-frequency signals. For the memory module, we adopt a pre-trained AR model to serve as the bank of linear filters, effectively incorporating the strengths of classical linear time series models into our architecture.

The memory module operates in the following manner: we first train an AR model from the input time series $\{x(n)\}$ to predict the next step via

\begin{equation}
    \hat{x}(n+1) = \sum_{i=0}^{p-1} a_i x(n - i),
\end{equation}
where $p$ is the AR order and $\{a_i\}_{i=0}^{p-1}$ are the learned AR coefficients. These coefficients are then extracted to define a set of fixed linear filters. At each time step $n$, a delay buffer forms the historical input vector $\{x(n - i)\}_{i=0}^{p-1}$, which is multiplied elementwise with the corresponding $\{a_i\}_{i=0}^{p-1}$ and passed to the subsequent KAN module. This structure is equivalent to setting the impulse response of the $i$-th filter in Fig.~\ref{img-theorem} as:

\begin{equation}
    h_i(n) = a_i \, \delta(n - i), \quad 0 \leq i \leq p-1,
\end{equation}
where $\delta(\cdot)$ is the Kronecker delta function.

To express the AR coefficients $\{a_i\}$ explicitly in terms of the time series $\{x(n)\}$, we can solve the Yule–Walker equations\cite{Yule_Walker1}\cite{Yule_Walker2}. Specifically, let $\mathbf{a} = [a_0, a_1, \dots, a_{p-1}]^\top$ be the coefficient vector, $\mathbf{r} = [r(1), r(2), \dots, r(p)]^\top$ the autocorrelation vector, and $\mathbf{R}$ the $p \times p$ autocorrelation matrix given by

\begin{equation}
    \mathbf{R} = \begin{bmatrix}
        r(0) & r(1) & \cdots & r(p-1) \\
        r(1) & r(0) & \cdots & r(p-2) \\
        \vdots & \vdots & \ddots & \vdots \\
        r(p-1) & r(p-2) & \cdots & r(0)
    \end{bmatrix},
\end{equation}
then the AR coefficients are computed via:

\begin{equation}
    \mathbf{a} = \mathbf{R}^{-1} \mathbf{r}.
\end{equation}

Here, the autocorrelation function \( r(i) \) is defined as

\begin{equation}
    r(i) = \mathbb{E}[x(n) \, x(n - i)],
\end{equation}

or, in practice, estimated from the empirical data as

\begin{equation}
    r(i) \approx \frac{1}{N - i} \sum_{n=i}^{N-1} x(n) \, x(n - i),
\end{equation}
where \( N \) is the total number of available samples.

This formulation reveals a key feature of our memory module: the filter weights $\{a_i\}$ are not fixed parameters, but are derived from the underlying data through statistical estimation. In contrast to static memory schemes such as tapped-delay lines\cite{Line_Memory} or gamma memory\cite{Gamma_Memory}, our data-driven design allows the memory module to adapt flexibly to the autocorrelation structure of different time series.

\subsection{Analysis of the AR Memory Module}

To further elucidate the advantage of the AR memory module, we provide a theoretical analysis demonstrating that it optimally preserves useful information while eliminating redundancy. Consider a general linear memory module with output:

\begin{equation}
y_i(n) = w_i x(n - i), \quad 0 \leq i \leq p-1,
\end{equation}
where $w_i$ are the weights.

We aim to maximize the total correlation between the memory outputs and the target $x(n+1)$, which represents the useful information captured:

\begin{equation}
\text{max} \; \sum_{i=0}^{p-1} \mathbb{E}[y_i(n) x(n+1)].
\end{equation}

However, this objective alone is insufficient, as it can be trivially maximized by arbitrarily increasing the magnitude of $w_i$, which would also amplify noise and irrelevant components. To prevent this and encourage the memory to focus on the most informative features, we introduce a constraint on the total output energy of the memory module:

\begin{equation}
\text{min}  \; \mathbb{E}\left[\left(\sum_{i=0}^{p-1} y_i(n)\right)^2\right].
\end{equation}

This constraint penalizes high-energy outputs, effectively forcing the memory to represent the target using a compact set of features and discard redundant information. We combine these two objectives into a single optimization goal:

\begin{equation}
L = \sum_{i=0}^{p-1} \mathbb{E}[y_i(n) x(n+1)] - \frac{1}{2} \mathbb{E}\left[\left(\sum_{i=0}^{p-1} y_i(n)\right)^2\right].
\end{equation}

To find the optimal weights that maximize $L$, we solve $\frac{\partial L}{\partial \mathbf{w}} = 0$ for $\mathbf{w} = [w_0, w_1, \dots, w_{p-1}]^\top$ gives:

\begin{equation}
\mathbf{w}^* = \mathbf{R}^{-1} \mathbf{r},
\end{equation}

which is exactly the solution for the AR coefficients. This result confirms that the AR memory module optimally balances the dual goals of preserving predictive information and minimizing redundancy, providing a principled foundation for its use in AR-KAN.

This adaptability endows AR-KAN with stronger generalization across diverse temporal patterns. The linear filters capture data-specific short-term dynamics, while the nonlinear KAN component models higher-order, nonlinear interactions. Together, they form a powerful hybrid that balances interpretability, efficiency, and expressiveness in time series forecasting.

\section{The Upper Bounds of Approximation Errors for KAN and AR-KAN}\label{Upper-Bounds}

In this section, we theoretically analyze and compare the approximation capabilities of the standard KAN and AR-KAN under simple structures. Specifically, we first employ the Jackson inequality\cite{jackson} to establish the upper bounds of their respective approximation errors. Subsequently, by incorporating the Minnesota Prior\cite{minnesota} and applying the Lyapunov Central Limit Theorem\cite{Lyapunov}, we probabilistically demonstrate that AR-KAN achieves a tighter error bound than the standard KAN.

\subsection{Theoretical Settings and Definitions}

\textbf{Assumption 1 (Input Space and Target Function):} 
Let the lagged input sequence of the time series be denoted as $\mathbf{x} = (x_1, x_2, \dots, x_p)^\top \in \Omega$. Without loss of generality, we assume the input is normalized such that the domain is the standard hypercube $\Omega = [-1, 1]^p$. Let the true underlying function generating the time series be $f: \Omega \to \mathbb{R}$.

\textbf{Assumption 2 (AR Weights and Smoothness Inductive Bias):} 
Assume the pre-trained AR weights form a diagonal matrix $\mathbf{W}_{AR} = \text{diag}(a_1, a_2, \dots, a_p)$. The time series generating function can be decomposed as:
\begin{equation}
    f(x_1, x_2, \dots, x_p) = \sum_{i=1}^p a_i x_i + \text{nonlinear terms}.
\end{equation}
In typical temporal forecasting scenarios, the linear AR components dominate, and the non-linear proportion is relatively small\cite{ARIMA-ANN}. Thus, we assume there exists a constant $c > 0$ such that:
\begin{equation}
    \sup_{\mathbf{x} \in \Omega} \left| \frac{\partial f}{\partial x_i} \right| \le c |a_i|, \quad \text{for } i = 1, 2, \dots, p.
\end{equation}
The physical intuition here is that a larger AR weight corresponds to a dimension where the function exhibits more drastic variations, whereas smaller weights indicate flatter dimensions.

\textbf{Assumption 3 (Mapping Relationships):} 
\begin{itemize}
    \item \textbf{Standard KAN:} Directly approximates the original function $f(\mathbf{x})$ over the domain $\Omega = [-1, 1]^p$.
    \item \textbf{AR-KAN:} Maps the input to a scaled space $\mathbf{z} = \mathbf{W}_{AR}\mathbf{x}$. The domain of $\mathbf{z}$ shrinks to $\tilde{\Omega} = \prod_{i=1}^p [-|a_i|, |a_i|]$. AR-KAN approximates a new target function $g(\mathbf{z})$ in this scaled space such that $g(\mathbf{W}_{AR}\mathbf{x}) = f(\mathbf{x})$, implying $g(\mathbf{z}) = f(\mathbf{W}_{AR}^{-1}\mathbf{z})$.
\end{itemize}
To ensure consistent model complexity, we assume both models use the same number of grid intervals $N$ for the B-splines in each dimension. The grid spacing $h_i$ depends on the domain length $L_i$, such that $h_i = \frac{L_i}{N}$.

\subsection{Derivation of the Approximation Error Bounds}

For a 1D function on a domain $D_i$ approximated by B-splines with $N$ intervals, the 1D Jackson inequality\cite{jackson} is given by:
\begin{equation}
    E_{1D}^2 \le C^2 h_i^2 \int_{D_i} \left( f'(u) \right)^2 du,
\end{equation}
where $C$ is a constant. To simplify the analysis, we consider a KAN with a $[p, 1]$ architecture. Therefore, the upper bound of its approximation error is proportional to the sum of the 1D errors across all dimensions:
\begin{equation}
    E_{KAN}^2 \le \sum_{i=1}^p C^2 h_i^2 \int_{D_i} \left( \frac{\partial f}{\partial x_i} \right)^2 dx_i.
\end{equation}

\subsubsection{Error Bound for Standard KAN ($E_{std}^2$)}
For standard KAN, the domain of the $i$-th dimension is $[-1, 1]$, yielding a length $L_i = 2$ and grid spacing $h_i = \frac{2}{N}$. Substituting this into the Jackson inequality gives:
\begin{equation}\label{eq_E_std}
    E_{std}^2 \le \frac{8C^2 c^2}{N^2} \sum_{i=1}^p |a_i|^2.
\end{equation}

\subsubsection{Error Bound for AR-KAN ($E_{AR}^2$)}
For AR-KAN, the domain of the $i$-th dimension is scaled to $[-|a_i|, |a_i|]$, with length $L_i = 2|a_i|$ and grid spacing $h_i = \frac{2|a_i|}{N}$. The error bound is:
\begin{equation}
    E_{AR}^2 \le \sum_{i=1}^p C^2 \left( \frac{2|a_i|}{N} \right)^2 \int_{-|a_i|}^{|a_i|} \left( \frac{\partial g}{\partial z_i} \right)^2 dz_i.
\end{equation}
By the chain rule, the partial derivative $\frac{\partial g}{\partial z_i}$ is:
\begin{equation}
    \left| \frac{\partial g}{\partial z_i} \right| = \left| \frac{\partial f}{\partial x_i} \cdot \frac{\partial x_i}{\partial z_i} \right| = \left| \frac{\partial f}{\partial x_i} \cdot \frac{1}{a_i} \right| \approx \frac{c|a_i|}{|a_i|} = c.
\end{equation}
Substituting this back, we obtain:
\begin{equation}\label{eq_E_ar}
    E_{AR}^2 \le \frac{8C^2 c^2}{N^2} \sum_{i=1}^p |a_i|^3.
\end{equation}

Comparing Equation (\ref{eq_E_std}) and (\ref{eq_E_ar}), discarding the common term $\frac{8C^2 c^2}{N^2}$, the core problem translates to determining the condition under which $\sum_{i=1}^p |a_i|^3 < \sum_{i=1}^p |a_i|^2$. If this holds, AR-KAN achieves a strictly smaller error bound than standard KAN under the same parameter budget.

\subsection{Probabilistic Analysis}

To evaluate this condition, we employ the Minnesota Prior framework \cite{minnesota}, assuming the AR weights $a_i$ are independent zero-mean normal variables with variance decaying strictly with $i$:
\begin{equation}
    a_i \sim \mathcal{N}(0, \sigma_i^2), \quad \text{where } \sigma_i^2 = \frac{k}{i^{2\alpha}}.
\end{equation}
Here, $\alpha > 0$ dictates the decay rate (typically $\alpha \ge 1$), and $k > 0$ is the variance scaling factor. 

Let $X_i = |a_i|^2 - |a_i|^3$. Our goal is to calculate the probability $P\left(\sum_{i=1}^p X_i > 0\right)$. Using the absolute moments of the normal distribution, we determine the expectation and variance of $X_i$:
\begin{align}
    \mathbb{E}[X_i] &= \sigma_i^2 - \sqrt{\frac{8}{\pi}} \sigma_i^3, \\
    Var[X_i] &= 2\sigma_i^4 - 12\sqrt{\frac{2}{\pi}}\sigma_i^5 + \left(15 - \frac{8}{\pi}\right)\sigma_i^6.
\end{align}

Since $X_i$ are independent and satisfy the Lyapunov condition\cite{Lyapunov}, the sum converges asymptotically to a normal distribution. By the Central Limit Theorem:
\begin{equation}
    P\left(\sum_{i=1}^p X_i > 0\right) \approx \Phi\left( \frac{\sum_{i=1}^p \mathbb{E}[X_i]}{\sqrt{\sum_{i=1}^p Var[X_i]}} \right),
\end{equation}
where $\Phi(\cdot)$ is the standard normal cumulative distribution function (CDF).

Introducing the Generalized Harmonic Number $H_p^{(m)} = \sum_{i=1}^p \frac{1}{i^m}$ (where $m$ denotes the exponent to avoid colliding with the lag order $p$), and noting that in practice $k \ll 1$, the higher-order terms become negligible. Dividing the numerator and denominator by $k$ simplifies the probability to:
\begin{equation}
    P\left(\sum_{i=1}^p X_i > 0\right) \approx \Phi\left( \frac{H_p^{(2\alpha)}}{\sqrt{2 H_p^{(4\alpha)}}} \right).
\end{equation}

\subsection{Asymptotic Behavior and Empirical Validation}

We apply the Euler-Maclaurin formula for the asymptotic expansion of the Generalized Harmonic Numbers:
\begin{equation}
    H_p^{(m)} = \zeta(m) - \frac{1}{(m-1)p^{m-1}} - \frac{1}{2p^m} + \mathcal{O}(p^{-m-1}),
\end{equation}
where $\zeta(m)$ is the Riemann zeta function. Applying this to both the numerator and the denominator, followed by a Taylor expansion, yields:
\begin{equation}
    \frac{H_p^{(2\alpha)}}{\sqrt{2 H_p^{(4\alpha)}}} = \frac{\zeta(2\alpha)}{\sqrt{2\zeta(4\alpha)}} - \frac{1}{\sqrt{2\zeta(4\alpha)}(2\alpha-1) p^{2\alpha-1}} + \mathcal{O}\left(\frac{1}{p^{2\alpha}}\right).
\end{equation}

\textbf{Conclusion:} As $p$ increases, the probability $P(E_{AR}^2 < E_{std}^2)$ approaches $\Phi\left( \frac{\zeta(2\alpha)}{\sqrt{2 \zeta(4\alpha)}} \right)$ from below at a rate of $\mathcal{O}\left( \frac{1}{p^{2\alpha-1}} \right)$. To explicitly illustrate this, we compute the convergence limits for several specific values of the decay rate $\alpha$, as presented in Table~\ref{tab-alpha-prob}.

\begin{table}[htbp]
    \centering
    \caption{The asymptotic probabilities for different $\alpha$.}
    \label{tab-alpha-prob}
    % --- 调整参数开始 ---
    \small
    \renewcommand{\arraystretch}{1.2}
    % --- 调整参数结束 ---
    \begin{tabular}{cc}
        \toprule
        $\alpha$ & $\Phi\left( \frac{\zeta(2\alpha)}{\sqrt{2 \zeta(4\alpha)}} \right)$ \\
        \midrule
        0.6 & $99.96\%$ \\
        0.8 & $93.27\%$ \\
        1.0 & $86.82\%$ \\
        1.2 & $83.09\%$ \\
        1.4 & $80.83\%$ \\
        \bottomrule
    \end{tabular}
\end{table}

\begin{figure}[htbp]
    \centering
    \includegraphics[width=0.98\linewidth]{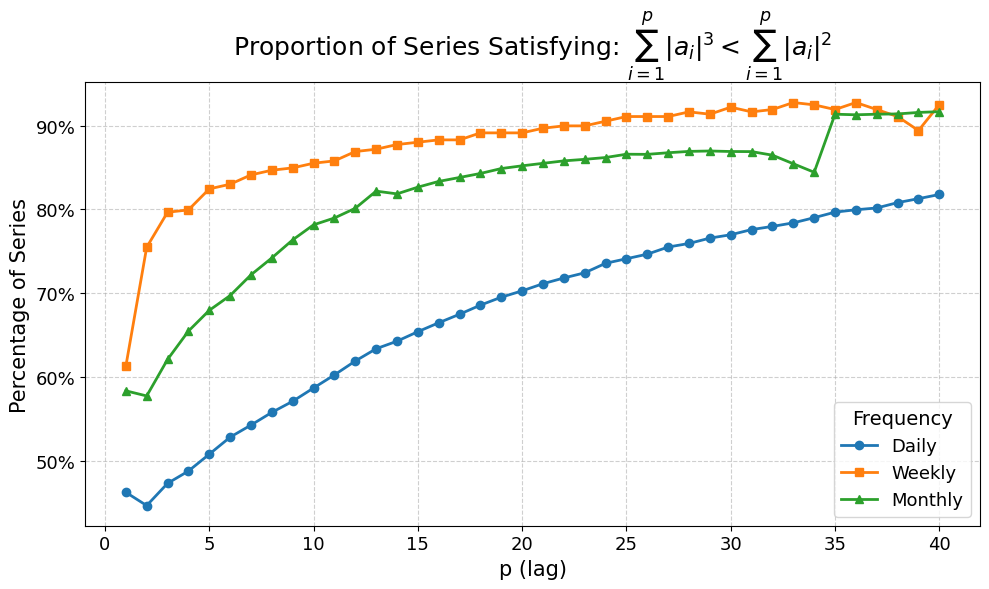}
    \caption{The proportion of series satisfying the condition $\sum_{i=1}^p |a_i|^3 < \sum_{i=1}^p |a_i|^2$ with respect to the number of lags $p$ on the M4 dataset.}
    \label{img-prob-p}
\end{figure}

This implies that under standard time series distributions, the error bound of AR-KAN is strictly smaller than that of standard KAN with high probability.

We further validated this condition on the M4 dataset\cite{m4}. As shown in Fig.~\ref{img-prob-p}, plotting the ratio of sequences satisfying $\sum_{i=1}^p |a_i|^3 < \sum_{i=1}^p |a_i|^2$ against the lag order $p \in [1, 40]$ confirms the theoretical derivation. The larger the model's lag order $p$, the easier it is to satisfy the condition, underscoring the intrinsic advantage of the AR-KAN architecture in real-world large-scale forecasting tasks.

\section{Experiments}\label{Experiments}
We conduct experiments in three parts to demonstrate both the effectiveness and generalizability of AR-KAN. First, we perform experiments on noisy almost-periodic functions to show that the majority of modern models fall short of traditional ARIMA models in terms of spectral analysis, while our AR-KAN achieves superior results. Then, we extend the evaluation to real-world datasets from Rdatasets \cite{rdatasets}, M3 \cite{m3}, and M4 \cite{m4} datasets. We observe that AR-KAN consistently achieves the best performance. Finally, we conduct ablation studies to demonstrate the necessity and effectiveness of each core component within our proposed architecture.

Additionally, for the sake of experimental efficiency, our architecture is built upon FastKAN\cite{fastkan} rather than the traditional KAN framework. By leveraging Radial Basis Functions (RBFs) for its implementation, FastKAN achieves faster computation while being theoretically proven to be equivalent to the original KAN representation.

\begin{table*}[htbp]
    \centering
    \caption{Model Performance Comparison on Noisy Almost Periodic Functions}
    \label{almost_periodic_table}

    \begin{tabular}{cc *{14}{c}}
        \toprule
        \multirow{2}{*}{lag} & \multirow{2}{*}{sigma} & \multicolumn{2}{c}{ARIMA} & \multicolumn{2}{c}{AR-KAN(ours)} & \multicolumn{2}{c}{TCN} & \multicolumn{2}{c}{Time-LLM} & \multicolumn{2}{c}{TimesNet} & \multicolumn{2}{c}{FAN} & \multicolumn{2}{c}{LSTM} \\
        \cmidrule(lr){3-4} \cmidrule(lr){5-6} \cmidrule(lr){7-8} \cmidrule(lr){9-10} \cmidrule(lr){11-12} \cmidrule(lr){13-14} \cmidrule(lr){15-16}
        & & $\text{R}^2$ & MAE & $\text{R}^2$ & MAE & $\text{R}^2$ & MAE & $\text{R}^2$ & MAE & $\text{R}^2$ & MAE & $\text{R}^2$ & MAE & $\text{R}^2$ & MAE \\
        \midrule
        
        \multirow{2}{*}{20} & 0.2 & 0.708 & 0.324 & \textcolor{red}{\textbf{0.895}} & \textcolor{red}{\textbf{0.208}} & \textcolor{blue}{\textbf{0.888}} & \textcolor{blue}{\textbf{0.221}} & \cellcolor{orange!15}0.304 & \cellcolor{orange!15}0.614 & \cellcolor{orange!15}0.154 & \cellcolor{orange!15}0.654 & \cellcolor{orange!15}0.276 & \cellcolor{orange!15}0.592 & \cellcolor{orange!15}-0.164 & \cellcolor{orange!15}0.709 \\
        
        & 0.4 & 0.633 & 0.457 & \textcolor{red}{\textbf{0.718}} & \textcolor{red}{\textbf{0.394}} & \textcolor{blue}{\textbf{0.652}} & \textcolor{blue}{\textbf{0.438}} & \cellcolor{orange!15}0.290 & \cellcolor{orange!15}0.663 & \cellcolor{orange!15}0.147 & \cellcolor{orange!15}0.740 & \cellcolor{orange!15}-0.719 & \cellcolor{orange!15}0.965 & \cellcolor{orange!15}-0.008 & \cellcolor{orange!15}0.704 \\
        \midrule
        
        \multirow{2}{*}{30} & 0.2 & 0.782 & 0.280 & \textcolor{red}{\textbf{0.913}} & \textcolor{red}{\textbf{0.191}} & \textcolor{blue}{\textbf{0.903}} & \textcolor{blue}{\textbf{0.201}} & \cellcolor{orange!15}0.759 & \cellcolor{orange!15}0.307 & \cellcolor{orange!15}0.738 & \cellcolor{orange!15}0.308 & 0.822 & 0.274 & \cellcolor{orange!15}0.575 & \cellcolor{orange!15}0.388 \\
        
        & 0.4 & 0.607 & 0.445 & \textcolor{red}{\textbf{0.749}} & \textcolor{red}{\textbf{0.368}} & \textcolor{blue}{\textbf{0.676}} & \textcolor{blue}{\textbf{0.413}} & \cellcolor{orange!15}0.538 & \cellcolor{orange!15}0.477 & \cellcolor{orange!15}0.506 & \cellcolor{orange!15}0.493 & \cellcolor{orange!15}-0.070 & \cellcolor{orange!15}0.652 & \cellcolor{orange!15}-0.079 & \cellcolor{orange!15}0.743 \\
        \midrule
        
        \multirow{2}{*}{40} & 0.2 & 0.868 & 0.217 & \textcolor{blue}{\textbf{0.907}} & \textcolor{blue}{\textbf{0.188}} & \textcolor{red}{\textbf{0.922}} & \textcolor{red}{\textbf{0.182}} & \cellcolor{orange!15}0.765 & \cellcolor{orange!15}0.308 & \cellcolor{orange!15}0.725 & \cellcolor{orange!15}0.329 & \cellcolor{orange!15}0.708 & \cellcolor{orange!15}0.341 & \cellcolor{orange!15}0.601 & \cellcolor{orange!15}0.397 \\
        
        & 0.4 & 0.567 & 0.457 & \textcolor{red}{\textbf{0.760}} & \textcolor{red}{\textbf{0.364}} & 0.606 & \textcolor{blue}{\textbf{0.429}} & 0.589 & \cellcolor{orange!15}0.461 & \cellcolor{orange!15}0.548 & \cellcolor{orange!15}0.482 & \textcolor{blue}{\textbf{0.651}} & 0.433 & \cellcolor{orange!15}0.559 & \cellcolor{orange!15}0.486 \\
        
        \bottomrule
    \end{tabular}

    \vspace{1ex}
    \raggedright
    \footnotesize
    \textit{Note:} $\text{R}^2$ represents the Coefficient of Determination (higher is better), and MAE represents the Mean Absolute Error (lower is better). \textbf{\textcolor{red}{Bold red}} and \textbf{\textcolor{blue}{bold blue}} values indicate the best and second-best results, respectively. Cells with a \colorbox{orange!15}{light orange background} indicate performance worse than the ARIMA baseline.

\end{table*}

\begin{table*}[htbp]
    \centering
    \caption{Model Performance Comparison on Real-World Datasets}
    \label{datasets_table}
    \resizebox{\textwidth}{!}{
    \begin{tabular}{cc *{21}{c}}
        \toprule
        \multirow{2}{*}{datasets} & \multirow{2}{*}{freq} & \multicolumn{3}{c}{ARIMA} & \multicolumn{3}{c}{AR-KAN(ours)} & \multicolumn{3}{c}{TCN} & \multicolumn{3}{c}{Time-LLM} & \multicolumn{3}{c}{TimesNet} & \multicolumn{3}{c}{FAN} & \multicolumn{3}{c}{LSTM} \\
        \cmidrule(lr){3-5} \cmidrule(lr){6-8} \cmidrule(lr){9-11} \cmidrule(lr){12-14} \cmidrule(lr){15-17} \cmidrule(lr){18-20} \cmidrule(lr){21-23}
        & & sMAPE & MASE & OWA & sMAPE & MASE & OWA & sMAPE & MASE & OWA & sMAPE & MASE & OWA & sMAPE & MASE & OWA & sMAPE & MASE & OWA & sMAPE & MASE & OWA \\
        \midrule
        
        \multirow{2}{*}{Rdatasets} & monthly & 6.96\% & 1.225 & 0.943 & \textcolor{red}{\textbf{5.01\%}} & \textcolor{red}{\textbf{0.788}} & \textcolor{red}{\textbf{0.470}} & 5.31\% & 0.888 & 0.634 & \textcolor{blue}{\textbf{5.18\%}} & \textcolor{blue}{\textbf{0.802}} & \textcolor{blue}{\textbf{0.528}} & 7.05\% & 1.232 & 0.751 & 5.46\% & 0.855 & 0.741 & 6.41\% & 1.096 & 0.694 \\
        
        & quarterly & 11.45\% & 2.293 & 0.634 & \textcolor{red}{\textbf{5.36\%}} & \textcolor{red}{\textbf{0.964}} & \textcolor{red}{\textbf{0.401}} & 6.82\% & 1.393 & \textcolor{blue}{\textbf{0.458}} & 7.45\% & 1.549 & 0.815 & 9.96\% & 2.056 & 0.578 & \textcolor{blue}{\textbf{6.49\%}} & \textcolor{blue}{\textbf{1.113}} & 0.745 & 9.88\% & 1.975 & 0.581 \\
        \midrule
        
        \multirow{2}{*}{M3} & monthly & 13.69\% & 0.901 & 0.776 & \textcolor{red}{\textbf{11.53\%}} & \textcolor{red}{\textbf{0.626}} & \textcolor{red}{\textbf{0.608}} & \textcolor{blue}{\textbf{11.84\%}} & 0.670 & \textcolor{blue}{\textbf{0.646}} & 12.61\% & \textcolor{blue}{\textbf{0.656}} & 0.741 & 13.30\% & 0.851 & 0.759 & 13.19\% & 0.753 & 0.712 & 13.28\% & 0.826 & 0.753 \\
        
        & quarterly & 11.60\% & 1.356 & 1.123 & \textcolor{red}{\textbf{7.16\%}} & \textcolor{red}{\textbf{0.810}} & \textcolor{red}{\textbf{0.721}} & \textcolor{blue}{\textbf{7.56\%}} & \textcolor{blue}{\textbf{0.851}} & \textcolor{blue}{\textbf{0.728}} & 10.04\% & 1.119 & 1.359 & 7.61\% & 0.914 & 0.761 & 7.92\% & 0.869 & 0.785 & 8.12\% & 0.937 & 0.798 \\
        \midrule
        
        \multirow{5}{*}{M4} & hourly & 15.17\% & 1.259 & 1.273 & 12.46\% & \textcolor{red}{\textbf{0.780}} & \textcolor{red}{\textbf{0.901}} & \textcolor{blue}{\textbf{11.07\%}} & 1.280 & 1.247 & 12.14\% & \textcolor{blue}{\textbf{0.845}} & 0.937 & 11.59\% & 0.860 & \textcolor{blue}{\textbf{0.904}} & 13.93\% & 0.907 & 0.940 & \textcolor{red}{\textbf{10.84\%}} & 0.953 & 0.986 \\
        
        & daily & 2.97\% & 3.184 & 1.711 & \textcolor{red}{\textbf{2.05\%}} & \textcolor{red}{\textbf{2.054}} & \textcolor{red}{\textbf{0.773}} & 2.56\% & 2.794 & 0.951 & 2.48\% & 2.585 & 0.923 & 3.62\% & 3.951 & 1.704 & 3.05\% & 3.263 & 1.270 & \textcolor{blue}{\textbf{2.41\%}} & \textcolor{blue}{\textbf{2.521}} & \textcolor{blue}{\textbf{0.877}} \\
        
        & weekly & 6.40\% & 0.424 & 0.651 & \textcolor{red}{\textbf{5.67\%}} & \textcolor{red}{\textbf{0.354}} & \textcolor{red}{\textbf{0.509}} & 6.30\% & 0.468 & 0.582 & 8.01\% & 0.592 & 0.848 & 7.82\% & 0.567 & 0.820 & 7.11\% & 0.449 & 0.609 & \textcolor{blue}{\textbf{6.28\%}} & \textcolor{blue}{\textbf{0.414}} & \textcolor{blue}{\textbf{0.565}} \\
        
        & monthly & \textcolor{blue}{\textbf{11.59\%}} & \textcolor{blue}{\textbf{0.838}} & 0.777 & \textcolor{red}{\textbf{10.87\%}} & \textcolor{red}{\textbf{0.727}} & \textcolor{red}{\textbf{0.748}} & 13.74\% & 0.902 & \textcolor{blue}{\textbf{0.753}} & 13.02\% & 0.965 & 0.909 & 12.97\% & 0.970 & 0.886 & 14.94\% & 1.048 & 1.033 & 12.63\% & 0.940 & 0.857 \\
        
        & quarterly & 10.15\% & 1.190 & 0.922 & \textcolor{red}{\textbf{8.85\%}} & \textcolor{red}{\textbf{0.975}} & \textcolor{red}{\textbf{0.734}} & \textcolor{blue}{\textbf{9.32\%}} & \textcolor{blue}{\textbf{1.075}} & \textcolor{blue}{\textbf{0.776}} & 10.10\% & 1.172 & 0.882 & 10.58\% & 1.244 & 0.919 & 12.83\% & 1.415 & 1.135 & 10.57\% & 1.317 & 0.933 \\
        
        \bottomrule
    \end{tabular}
    }
    
    \vspace{1ex}
    \raggedright
    \footnotesize
    \textit{Note:} sMAPE represents the Symmetric Mean Absolute Percentage Error, MASE represents the Mean Absolute Scaled Error, and OWA represents the Overall Weighted Average. For all three metrics, lower values indicate better forecasting performance. \textbf{\textcolor{red}{Bold red}} and \textbf{\textcolor{blue}{bold blue}} values indicate the best and second-best results, respectively. The metrics reported above are calculated based on the best parameters obtained through grid search.
\end{table*}

\begin{figure}[htbp]
    \centering
    \includegraphics[width=0.98\linewidth]{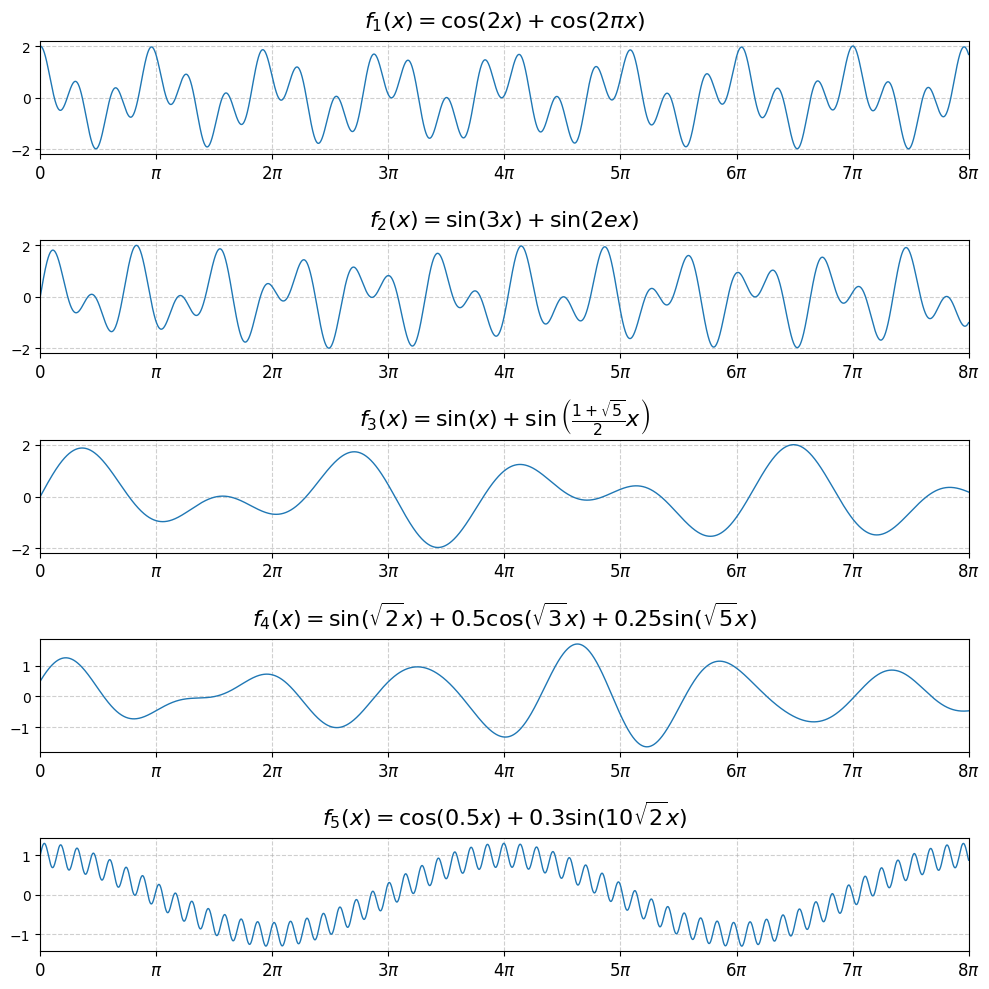}
    \caption{Visualization of the five generated almost-periodic functions.}
    \label{almost_periodic_figure}
\end{figure}

\subsection{Noisy Almost Periodic Functions}
We construct noisy almost-periodic functions by superimposing multiple trigonometric waves with incommensurate frequencies and adding Gaussian noise. The waveforms of these five almost-periodic functions are illustrated in Fig. \ref{almost_periodic_figure}.

where the noise is sampled from a zero-mean Gaussian distribution with variance $\sigma^2$. Almost-periodic functions like this are of particular significance in the development of harmonic analysis, and they form the basis of generalized harmonic analysis (GHA) as formulated by Wiener\cite{wienerGHA}.

Specifically, for each function, we uniformly sample 500 points over the interval $[0, 8\pi]$. We set the noise level $\sigma$ to 0.2 and 0.4, and compare the performance of ARIMA with six neural models. The dataset is split into training and testing sets with an 80/20 ratio: the first 80\% of the sequence is used for training, while the remaining 20\% is reserved for testing. The results shown in TABLE~\ref{almost_periodic_table} represent the average performance across all five functions.

It is evident that, among the existing neural network models, with the exception of TCN, the majority of modern baselines struggle significantly and perform even worse than the traditional ARIMA model (highlighted by the light orange cells). In contrast, our proposed AR-KAN consistently achieves the optimal or near-optimal performance across various lag orders and noise levels. This demonstrates that while modern deep learning architectures possess powerful nonlinear representation capabilities, they lack the appropriate spectral learning and representation capacities to inherently capture almost-periodic dynamics with incommensurate frequencies. By successfully integrating the autoregressive prior, AR-KAN effectively overcomes this limitation, proving that the fusion of classical temporal memory and advanced nonlinear mapping is highly effective for complex spectral analysis.

\begin{table*}[htbp]
    \centering
    \caption{Ablation Studies on Noisy Almost Periodic Functions}
        \label{ablation_studies_almost_periodic_table}
    \begin{tabular}{cc cccccccccc}
        \toprule
        \multirow{2}{*}{lag} & \multirow{2}{*}{sigma} & \multicolumn{2}{c}{ARIMA} & \multicolumn{2}{c}{AR-KAN(ours)} & \multicolumn{2}{c}{AR-MLP} & \multicolumn{2}{c}{KAN} & \multicolumn{2}{c}{MLP} \\
        \cmidrule(lr){3-4} \cmidrule(lr){5-6} \cmidrule(lr){7-8} \cmidrule(lr){9-10} \cmidrule(lr){11-12}
        & & $\text{R}^2$ & MAE & $\text{R}^2$ & MAE & $\text{R}^2$ & MAE & $\text{R}^2$ & MAE & $\text{R}^2$ & MAE \\
        \midrule
        \multirow{2}{*}{20} 
        & 0.2 & 0.708 & 0.324 & \textcolor{blue}{\textbf{0.895}} & \textcolor{red}{\textbf{0.208}} & \textcolor{red}{\textbf{0.902}} & \textcolor{blue}{\textbf{0.212}} & \cellcolor{orange!15}-0.623 & \cellcolor{orange!15}0.683 & \cellcolor{orange!15}0.366 & \cellcolor{orange!15}0.439 \\
        & 0.4 & 0.633 & 0.457 & \textcolor{red}{\textbf{0.718}} & \textcolor{red}{\textbf{0.394}} & \textcolor{blue}{\textbf{0.651}} & \textcolor{blue}{\textbf{0.440}} & \cellcolor{orange!15}-0.748 & \cellcolor{orange!15}0.986 & \cellcolor{orange!15}0.174 & \cellcolor{orange!15}0.704 \\
        \midrule
        \multirow{2}{*}{30} 
        & 0.2 & 0.782 & 0.280 & \textcolor{red}{\textbf{0.913}} & \textcolor{blue}{\textbf{0.191}} & \textcolor{blue}{\textbf{0.908}} & \textcolor{red}{\textbf{0.189}} & \cellcolor{orange!15}0.422 & \cellcolor{orange!15}0.397 & \cellcolor{orange!15}0.273 & \cellcolor{orange!15}0.519 \\
        & 0.4 & 0.607 & 0.445 & \textcolor{red}{\textbf{0.749}} & \textcolor{red}{\textbf{0.368}} & \textcolor{blue}{\textbf{0.666}} & \textcolor{blue}{\textbf{0.420}} & \cellcolor{orange!15}-0.307 & \cellcolor{orange!15}0.748 & \cellcolor{orange!15}0.118 & \cellcolor{orange!15}0.688 \\
        \midrule
        \multirow{2}{*}{40} 
        & 0.2 & 0.868 & 0.217 & \textcolor{blue}{\textbf{0.907}} & \textcolor{blue}{\textbf{0.188}} & \textcolor{red}{\textbf{0.916}} & \textcolor{red}{\textbf{0.187}} & \cellcolor{orange!15}0.725 & \cellcolor{orange!15}0.326 & \cellcolor{orange!15}0.438 & \cellcolor{orange!15}0.411 \\
        & 0.4 & 0.567 & \textcolor{blue}{\textbf{0.457}} & \textcolor{red}{\textbf{0.760}} & \textcolor{red}{\textbf{0.364}} & \textcolor{blue}{\textbf{0.581}} & \cellcolor{orange!15}0.465 & \cellcolor{orange!15}-0.050 & \cellcolor{orange!15}0.629 & \cellcolor{orange!15}0.316 & \cellcolor{orange!15}0.565 \\
        \bottomrule
    \end{tabular}

    \vspace{1ex}
    \raggedright
    \footnotesize
    \textit{Note:} $\text{R}^2$ represents the Coefficient of Determination (higher is better), and MAE represents the Mean Absolute Error (lower is better). \textbf{\textcolor{red}{Bold red}} and \textbf{\textcolor{blue}{bold blue}} values indicate the best and second-best results, respectively. Cells with a \colorbox{orange!15}{light orange background} indicate performance worse than the ARIMA baseline.
\end{table*}

\begin{table*}[htbp]
    \centering
    \caption{Ablation Studies on Real-World Datasets}
    \label{ablation_studies_datasets_table}
    \resizebox{\textwidth}{!}{
        \begin{tabular}{cc ccc ccc ccc ccc ccc}
            \toprule
            \multirow{2}{*}{datasets} & \multirow{2}{*}{freq} & \multicolumn{3}{c}{ARIMA} & \multicolumn{3}{c}{AR-KAN(ours)} & \multicolumn{3}{c}{AR-MLP} & \multicolumn{3}{c}{KAN} & \multicolumn{3}{c}{MLP} \\
            \cmidrule(lr){3-5} \cmidrule(lr){6-8} \cmidrule(lr){9-11} \cmidrule(lr){12-14} \cmidrule(lr){15-17}
            & & sMAPE & MASE & OWA & sMAPE & MASE & OWA & sMAPE & MASE & OWA & sMAPE & MASE & OWA & sMAPE & MASE & OWA \\
            \midrule
            \multirow{2}{*}{Rdatasets} 
            & monthly & 6.96\% & 1.225 & 0.943 & \textcolor{red}{\textbf{5.01\%}} & \textcolor{red}{\textbf{0.788}} & \textcolor{red}{\textbf{0.470}} & \textcolor{blue}{\textbf{5.12\%}} & \textcolor{blue}{\textbf{0.793}} & \textcolor{blue}{\textbf{0.489}} & 6.98\% & 1.208 & 0.609 & 5.32\% & 0.794 & 0.520 \\
            & quarterly & 11.45\% & 2.293 & 0.634 & \textcolor{red}{\textbf{5.36\%}} & \textcolor{blue}{\textbf{0.964}} & \textcolor{red}{\textbf{0.401}} & \textcolor{blue}{\textbf{5.43\%}} & \textcolor{red}{\textbf{0.918}} & \textcolor{blue}{\textbf{0.415}} & 7.77\% & 1.344 & 0.526 & 6.46\% & 1.065 & 0.472 \\
            \midrule
            \multirow{2}{*}{M3} 
            & monthly & 13.69\% & 0.901 & 0.776 & \textcolor{red}{\textbf{11.53\%}} & \textcolor{red}{\textbf{0.626}} & \textcolor{red}{\textbf{0.608}} & \textcolor{blue}{\textbf{12.64\%}} & \textcolor{blue}{\textbf{0.687}} & \textcolor{blue}{\textbf{0.666}} & 12.74\% & 0.776 & 0.718 & 13.53\% & 0.720 & 0.714 \\
            & quarterly & 11.60\% & 1.356 & 1.123 & \textcolor{blue}{\textbf{7.16\%}} & \textcolor{red}{\textbf{0.810}} & \textcolor{red}{\textbf{0.721}} & 9.27\% & 1.042 & 0.961 & \textcolor{red}{\textbf{7.08\%}} & \textcolor{blue}{\textbf{0.827}} & \textcolor{blue}{\textbf{0.725}} & 8.85\% & 1.002 & 0.951 \\
            \midrule
            \multirow{5}{*}{M4} 
            & hourly & 15.17\% & 1.259 & 1.273 & \textcolor{red}{\textbf{12.46\%}} & \textcolor{red}{\textbf{0.780}} & \textcolor{red}{\textbf{0.901}} & 13.77\% & 1.626 & 1.208 & \textcolor{blue}{\textbf{12.67\%}} & 0.857 & \textcolor{blue}{\textbf{0.950}} & 15.76\% & \textcolor{blue}{\textbf{0.789}} & 1.082 \\
            & daily & 2.97\% & 3.184 & 1.711 & \textcolor{red}{\textbf{2.05\%}} & \textcolor{red}{\textbf{2.054}} & \textcolor{red}{\textbf{0.773}} & 2.70\% & 2.857 & \textcolor{blue}{\textbf{0.957}} & \textcolor{blue}{\textbf{2.51\%}} & \textcolor{blue}{\textbf{2.605}} & 0.987 & 2.83\% & 2.983 & 1.099 \\
            & weekly & 6.40\% & 0.424 & 0.651 & \textcolor{red}{\textbf{5.67\%}} & \textcolor{red}{\textbf{0.354}} & \textcolor{red}{\textbf{0.509}} & 6.79\% & 0.429 & 0.595 & \textcolor{blue}{\textbf{5.84\%}} & \textcolor{blue}{\textbf{0.368}} & \textcolor{blue}{\textbf{0.514}} & 6.12\% & 0.396 & 0.556 \\
            & monthly & \textcolor{blue}{\textbf{11.59\%}} & \textcolor{blue}{\textbf{0.838}} & \textcolor{blue}{\textbf{0.777}} & \textcolor{red}{\textbf{10.87\%}} & \textcolor{red}{\textbf{0.727}} & \textcolor{red}{\textbf{0.748}} & 13.73\% & 1.103 & 0.934 & 12.12\% & 0.859 & 0.802 & 14.81\% & 1.054 & 1.011 \\
            & quarterly & 10.15\% & 1.190 & 0.922 & \textcolor{red}{\textbf{8.85\%}} & \textcolor{red}{\textbf{0.975}} & \textcolor{red}{\textbf{0.734}} & 11.89\% & 1.422 & 1.092 & \textcolor{blue}{\textbf{9.70\%}} & \textcolor{blue}{\textbf{1.096}} & \textcolor{blue}{\textbf{0.840}} & 12.69\% & 1.442 & 1.156 \\
            \bottomrule
        \end{tabular}
    }
    
    \vspace{1ex}
    \raggedright
    \footnotesize
    \textit{Note:} sMAPE represents the Symmetric Mean Absolute Percentage Error, MASE represents the Mean Absolute Scaled Error, and OWA represents the Overall Weighted Average. For all three metrics, lower values indicate better forecasting performance. \textbf{\textcolor{red}{Bold red}} and \textbf{\textcolor{blue}{bold blue}} values indicate the best and second-best results, respectively. The metrics reported above are calculated based on the best parameters obtained through grid search.
\end{table*}

\subsection{Real-World Datasets}

In this subsection, we evaluate the practical applicability of our AR-KAN model using real-world time series. Specifically, we employ three well-known datasets: Rdatasets, M3, and M4 datasets. These datasets are widely recognized as standard benchmarks in the field of time series forecasting, encompassing a broad spectrum of domains—such as finance, macroeconomics, industry, and demographics—and exhibiting diverse, complex temporal dynamics. The experimental results are summarized in TABLE~\ref{datasets_table}. The reported metrics represent the optimal performance of each model, obtained after a comprehensive grid search over the hyperparameter space. Detailed experimental setups and configurations are provided in APPENDIX \ref{model_settings}.

It is observed that AR-KAN demonstrates remarkable and consistent superiority across all evaluated real-world datasets and sampling frequencies, comprehensively outperforming both classical statistical methods and modern deep learning baselines. Notably, our model consistently achieves optimal performance on the Overall Weighted Average (OWA)—a critical metric for comprehensive forecasting evaluation—establishing a significant gap between AR-KAN and highly competitive neural architectures like TCN. Furthermore, the macro-level results indicate that heavily parameterized models or repurposed large language models, such as Time-LLM and TimesNet, do not inherently guarantee better forecasting accuracy. Instead, AR-KAN's success across diverse temporal resolutions, ranging from highly volatile high-frequency data to macro-trend low-frequency data, validates our core hypothesis: explicitly grounding a model with classical autoregressive priors while leveraging the dynamic, non-linear representation capabilities of KAN provides a fundamentally more robust, efficient, and generalizable solution for complex real-world time series forecasting.

\subsection{Ablation Studies}
Finally, to validate the necessity and individual contributions of each core component within our proposed architecture, we conduct comprehensive ablation studies. Specifically, we investigate the impact of integrating the Autoregressive (AR) memory module, while concurrently comparing the representational capabilities of KAN against MLP. The variants evaluated in this ablation analysis include ARIMA, AR-KAN (our full model), AR-MLP, standard KAN, and standard MLP. All ablation experiments strictly follow the identical settings and evaluation protocols established in the previous two sections. The results of these comparisons are presented in TABLE~\ref{ablation_studies_almost_periodic_table} and TABLE~\ref{ablation_studies_datasets_table}.

As indicated by the table, standalone MLP and KAN architectures struggle significantly, exhibiting severely degraded forecasting performance. However, upon integrating the AR memory module, the performance of both networks improves drastically. This compelling contrast demonstrates that pure feed-forward neural networks fundamentally lack the proper spectral representation capacity to inherently capture long-term temporal dependencies and incommensurate periodic dynamics. The AR module successfully bridges this gap by providing a crucial linear temporal baseline. Moreover, while AR-MLP and AR-KAN yield highly comparable results in this specific synthetic scenario, AR-KAN consistently maintains a distinct edge, demonstrating a more precise nonlinear fitting capability.
﻿
Turning to the real-world datasets, the inherent advantages of the KAN architecture become highly prominent. AR-KAN decisively secures the vast majority of first-place rankings across diverse domains and frequencies, while the standalone KAN model captures approximately half of the second-place positions. In stark contrast, both AR-MLP and standard MLP exhibit merely mediocre performance. This significant performance gap highlights the fundamental theoretical advantages of KAN over traditional MLPs in terms of spectral bias and frequency analysis. As previously discussed, KANs are largely resilient to the spectral bias that inherently plagues standard MLPs. This crucial advantage enables KAN to uniformly and efficiently learn across a broad spectrum of frequencies, rather than being biased towards merely fitting dominant low-frequency signals. Consequently, KAN can adeptly capture the intricate, high-frequency fluctuations ubiquitous in real-world time series. By coupling this with the AR module—which reliably anchors the overall low-frequency temporal trends—AR-KAN forms a powerful synergy that achieves superior forecasting accuracy across highly complex and varied temporal dynamics.

\section{Conclusion}\label{Conclusion}
In this paper, we proposed the Autoregressive-Weight-Enhanced Kolmogorov-Arnold Network (AR-KAN) to address the inherent limitations of modern neural architectures in capturing complex spectral structures, particularly almost-periodic signals with incommensurate frequencies. Grounded in the Universal Myopic Mapping Theorem, AR-KAN strategically fuses a data-driven AR memory module with the flexible, frequency-unbiased nonlinear expressiveness of KANs. Theoretically, we proved that the AR module optimally preserves essential predictive information while minimizing redundancy. Furthermore, we established the upper bounds of the approximation errors for KAN and AR-KAN under simple structures, and proved that under the Minnesota Prior framework, the error bound of AR-KAN is strictly smaller than that of a standard KAN with a probability that increases with the lag order $p$. Extensive empirical evaluations across synthetic almost-periodic functions and diverse real-world benchmarks—including Rdatasets, M3, and M4—alongside comprehensive ablation studies, consistently validate our theoretical findings. By seamlessly integrating the strong spectral bias of classical autoregressive priors with the advanced nonlinear representation capabilities of KANs, AR-KAN emerges as a highly robust, interpretable, and theoretically grounded framework for general time series forecasting.

\appendices

\section{Model Architecture and Configuration}\label{model_settings}

\begin{table}[htbp]
    \centering
    \renewcommand{\arraystretch}{1.5} % 增加行高
    % 注意：如果有 n 个 X 列，\hsize 的系数总和必须正好等于 n
    \begin{tabularx}{\linewidth}{
        >{\hsize=0.4\hsize\centering\arraybackslash}X 
        >{\hsize=1.6\hsize\centering\arraybackslash}X
    }
        \toprule
        \textbf{Models} & \textbf{Architecture and Configuration} \\
        \midrule
        
        ARIMA & 
        $p=20$, $d \in \{0,1\}$, $q \in \{0,1,2\}$ \\
        
        MLP & 
        lag=20, hidden\_dim $\in \{50, 100, 150, 200, 250, 300\}$, batch=64, lr=0.001, epochs=100 \\
        
        KAN & 
        FastKAN, lag=20, hidden\_dim $\in \{50, 100, 150\}$, grids $\in \{3,5\}$, batch=64, lr=0.001, epochs=100 \\
        
        LSTM & 
        lag=20, hidden\_dim $\in \{32, 64, 128\}$, num\_layers $\in \{1,2\}$, batch=64, lr=0.001, epochs=100 \\
        
        FAN & 
        lag=20, hidden\_dim $\in \{32, 64, 128\}$, num\_layers $\in \{2,4\}$, batch=64, lr=0.001, epochs=100 \\
        
        TimesNet & 
        lag=20, top\_k=3, num\_kernels=6, hidden\_dim $\in \{16, 32, 64\}$, num\_layers $\in \{2,4\}$, batch=64, lr=0.001, epochs=100 \\
        
        TCN & 
        lag=20, kernel\_size=3, hidden\_dim $\in \{16, 32, 64\}$, num\_layers $\in \{2,4\}$, batch=64, lr=0.001, epochs=100 \\
        
        Time-LLM & 
        LLaMA-7B \\
        
        \bottomrule
    \end{tabularx}
\end{table}

\section{Evaluation Metrics}\label{appendix_metrics}

In the Noisy Almost Periodic Functions experiments, we evaluate the model forecasting performance using the Coefficient of Determination ($R^2$) and the Mean Absolute Error (MAE). For the Real-World Datasets experiments, to ensure a comprehensive and standardized evaluation, we uniformly adopt the official metrics utilized in the M4 competition \cite{m4}: Symmetric Mean Absolute Percentage Error (sMAPE), Mean Absolute Scaled Error (MASE), and the Overall Weighted Average (OWA). 

Given a sequence of true values $\mathbf{y}$ and predicted values $\mathbf{\hat{y}}$ over a forecasting horizon $h$, the definitions and mathematical formulations of these metrics are detailed as follows:

\textbf{Mean Absolute Error (MAE)} measures the average magnitude of the absolute differences between the predicted and true values, reflecting the overall prediction scale:
\begin{equation}
    \text{MAE} = \frac{1}{h} \sum_{t=1}^{h} |y_t - \hat{y}_t|
\end{equation}

\textbf{Coefficient of Determination ($R^2$)} evaluates the proportion of the variance in the dependent variable that is predictable from the model. A higher $R^2$ (closer to 1) indicates a better fit:
\begin{equation}
    R^2 = 1 - \frac{\sum_{t=1}^{h} (y_t - \hat{y}_t)^2}{\sum_{t=1}^{h} y_t^2}
\end{equation}

\textbf{Symmetric Mean Absolute Percentage Error (sMAPE)} is a relative metric that computes the percentage error based on the average of the true and predicted values, effectively handling values close to zero:
\begin{equation}
    \text{sMAPE} = \frac{1}{h} \sum_{t=1}^{h} \frac{2 |y_t - \hat{y}_t|}{|y_t| + |\hat{y}_t|} \times 100\%
\end{equation}

\textbf{Mean Absolute Scaled Error (MASE)} is a scale-independent error metric. It compares the MAE of the forecasts with the MAE of a naive random walk forecast computed on the historical training data. It is highly suitable for datasets with diverse scales:
\begin{equation}
    \text{MASE} = \frac{\frac{1}{h} \sum_{t=1}^{h} |y_t - \hat{y}_t|}{\frac{1}{n-m} \sum_{j=m+1}^{n} |y_j - y_{j-m}|}
\end{equation}

\textbf{Overall Weighted Average (OWA)} is a comprehensive metric specifically proposed in the M4 competition. It calculates the average of the relative sMAPE and MASE compared to the standard Naive2 baseline model, providing a robust macro-level evaluation:
\begin{equation}
    \text{OWA} = \frac{1}{2} \left( \frac{\text{sMAPE}}{\text{sMAPE}_{\text{Naive2}}} + \frac{\text{MASE}}{\text{MASE}_{\text{Naive2}}} \right)
\end{equation}

\textbf{Parameter Definitions:}
\begin{itemize}
    \item $y_t$ and $\hat{y}_t$ denote the actual ground-truth value and the predicted value at time step $t$, respectively.
    \item $h$ is the total length of the forecasting horizon (i.e., the number of predicted points).
    \item $n$ is the total length of the historical training time series.
    \item $m$ represents the seasonal frequency or periodicity of the time series as defined in the M4 competition (e.g., $m=4$ for quarterly, $m=12$ for monthly, $m=52$ for weekly, $m=7$ for daily, and $m=24$ for hourly data).
    \item $\text{sMAPE}_{\text{Naive2}}$ and $\text{MASE}_{\text{Naive2}}$ are the corresponding metric scores achieved by the standard Naive2 baseline model on the same forecasting task.
\end{itemize}

\bibliographystyle{IEEEtran}

\bibliography{reference.bib}

\end{document}